\documentclass[conference]{IEEEtran}
\IEEEoverridecommandlockouts

\usepackage{subcaption}
\usepackage{float}
\usepackage{cite}
\usepackage{amsmath,amssymb,amsfonts}
\usepackage{algorithmic}
\usepackage{tabularx}
\usepackage{textcomp}
\usepackage{xcolor}
\usepackage{booktabs}
\usepackage{multirow}

\def\BibTeX{{\rm B\kern-.05em{\sc i\kern-.025em b}\kern-.08em
    T\kern-.1667em\lower.7ex\hbox{E}\kern-.125emX}}

\ifCLASSINFOpdf
  \usepackage[pdftex]{graphicx}
\else
\fi

\begin{document}

% \title{An Empirical Investigation into Human-Model Alignment on Multiple-Human-Annotation Datasets }

\title{Does Model Uncertainty Track Human Ambiguity? Evidence from Multi-Annotator Vision Benchmarks}
% \author{\IEEEauthorblockN{Manya Singh}, \IEEEauthorblockN{Arjun Pakrashi}}% <-this % stops a space

% \author{
% \IEEEauthorblockN{Manya Singh \AND Arjun Pakrashi}
% \IEEEauthorblockA{
% School of Computer Science\\
% University College Dublin\\
% Dublin, Ireland\\
% manya.singh@ucdconnect.ie, arjun.pakrashi@ucd.ie
% }
% }

\author{
\IEEEauthorblockN{Manya Singh and Arjun Pakrashi}
\IEEEauthorblockA{
School of Computer Science, University College Dublin, Ireland\\
manya.singh@ucdconnect.ie, arjun.pakrashi@ucd.ie
}
}

% \markboth{Journal of \LaTeX\ Class Files,~Vol.~14, No.~8, August~2015}%
% {Shell \MakeLowercase{\textit{et al.}}: Bare Demo of IEEEtran.cls for IEEE Journals}

\maketitle

\begin{abstract}
Human-model alignment is critical for trustworthy AI-assisted decision-making systems. Yet, most work evaluates model predictions against single ground-truth labels, overlooking that humans themselves often disagree on labels, a signal of genuine ambiguity. We investigate whether models struggle on the same instances that humans find difficult. We measure this on two vision datasets (\texttt{FER+} and \texttt{CIFAR-10H}) where multiple human annotations per image capture human disagreement patterns. We evaluate eight pretrained models across three architectures (\texttt{ResNet}, \texttt{EfficientNet}, \texttt{MobileNetV3}) in two parts: first, whether model uncertainty (softmax confidence, entropy) correlates with human disagreement, and second, whether predictive multiplicity measures (inter-model disagreement, Jensen-Shannon divergence) do. We find that it does not: alignment is weak in both dimensions. At the discrete label level, 50.4\% of \texttt{CIFAR-10H} images and 33.5\% of \texttt{FER+} images receive multiple valid classifications from humans, while the models converge on only one. These instances represent a critical failure case where humans perceive ambiguity and would request expert review, yet models decide confidently. At the continuous score level, single-model uncertainty correlates weakly with human disagreement ($\rho = 0.24$--$0.55$), and predictive multiplicity provides only modest improvement. Widely-used uncertainty quantification methods do not reliably identify instances humans find ambiguous. Model uncertainty should not be treated as a trustworthy signal by default for decision-making in high-stakes scenarios.
\end{abstract}


\section{Introduction}
\begin{figure}[t]
  \centering
  \includegraphics[width=\columnwidth]{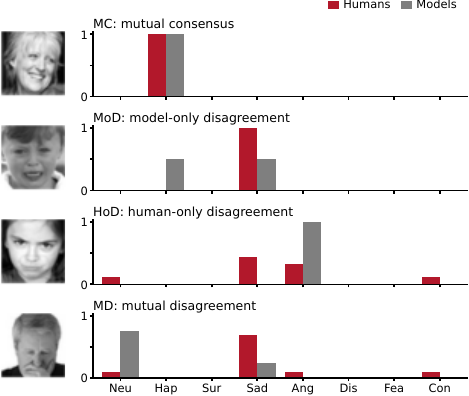}
  \caption{Representative \texttt{FER+} instances for the four human--model disagreement profiles. Each row shows the test image with the distribution of its $10$ human votes (red) and of the $8$ model predictions (grey) over the emotion classes: Neu(tral), Hap(piness), Sur(prise), Sad(ness), Ang(er), Dis(gust), Fea(r), Con(tempt).}
  \label{fig:quadrants}
\end{figure}

% #para 1 - what is the context?

Machine learning (ML) models are evaluated by measuring the alignment between \textit{model predictions} and \textit{human judgments}. When both are collapsed to single labels, a majority vote and a predicted class, predictive accuracy directly measures alignment through a point-to-point estimate comparison. But when both remain \textit{distributions}, the comparison becomes fundamentally different \cite{pavlick-kwiatkowski-2019-inherent}. Model uncertainty captures the distribution of the model's confidence, while human uncertainty captures disagreement among human annotators. Without validating model uncertainty against the full distribution of human annotations through a distribution-to-distribution comparison, we cannot assess its reliability. Performing such distribution-to-distribution comparisons faces a critical barrier. Most datasets only provide a single ground-truth label per instance, typically a majority vote from human annotators \cite{davani2022dealing, lan2025mind}. This structure enables point-to-point evaluation (accuracy) but makes distribution-to-distribution (uncertainty) validation difficult.

% #para 2 - so what research problem does this create?
However, uncertainty-based decision signals have still been deployed in high-stakes ML systems. In active learning, models select the most uncertain examples for annotation, directing labeling budgets \cite{settles2009active, lewis1994sequential, geifman2017deep}. In selective classification or abstention \cite{geifman2019selectivenet}, models reject inputs when uncertainty is high to avoid unsafe decisions. In bio-signal applications \cite{de2026uncertainty}, uncertainty measures guide clinical decision support, where clinicians are only shown samples that the model struggles to predict with high certainty. In autonomous driving vehicles \cite{michelmore2018evaluating}, model uncertainty (high or low) is used as a binary signal to decide whether the car needs to steer away from a crash. All these systems assume that model uncertainty aligns with human difficulty. We state this assumption as a hypothesis: \emph{instances on which human annotators divide should be instances on which models hesitate, through low confidence, high entropy, or conflicting predictions.} Without systematic validation against human disagreement distributions, this hypothesis has remained untested for modern vision models.

% #para 3 - how do you solve this problem?
%In this work, our main objective is to quantify the alignment between human disagreement (divergence among multiple annotators) and model-side uncertainty measures on two benchmark vision datasets, \texttt{FER+} and \texttt{CIFAR-10H}, where multiple human annotations are available. 
In this work, we aim to quantify the alignment between human disagreement (disagreements between multiple annotators) and model uncertainty measures on benchmark vision datasets (i.e., \texttt{FER+} and \texttt{CIFAR-10H}).
Unlike previous works, which compare a single model's uncertainty estimates to human annotator disagreement~\cite{lan2025mind,mendes2025uncertainty}, we additionally consider, to the best of our knowledge for the first time, predictive multiplicity~\cite{marx2020predictive} among eight vision models across three model families (\texttt{ResNet}, \texttt{EfficientNet}, \texttt{MobileNetV3}). We find that model uncertainty correlates only weakly to moderately with human disagreement ($\rho = 0.24$--$0.55$), and that predictive multiplicity offers modest improvement. The discrete counterpart of this misalignment is stark: $50.4\%$ of \texttt{CIFAR-10H} and $33.5\%$ of \texttt{FER+} test instances receive a unanimous prediction from all eight models while human annotators divide (Fig.~\ref{fig:quadrants}). The two datasets further differ in whether fine-tuning labels derive from majority votes over multiple annotators (\texttt{FER+}) or from single labels (\texttt{CIFAR-10H}), letting us observe how label provenance relates to alignment. We characterise where this misalignment concentrates and what it implies for trustworthy ML deployment.

\begin{table*}[t]
\caption{Definitions of key terms used in this work.}
\label{tab:key_terms}
\centering
\scriptsize
\renewcommand{\arraystretch}{1.25}
\begin{tabularx}{\textwidth}{l X}
\toprule
Term & Definition \\
\midrule
Human disagreement & Uncertainty arising from disagreement among human annotators when assigning labels to the same instance, as defined in~\cite{mendes2025uncertainty,lan2025mind} (there termed \emph{human uncertainty}). It captures the inherent ambiguity or subjectivity of the ground-truth label. \\
Model uncertainty & Uncertainty in a single model's prediction, typically
quantified using predictive entropy or another uncertainty score derived from its predicted class probabilities~\cite{gawlikowski2023survey,hullermeier2021aleatoric}. It captures how confident a model is about its prediction. \\
Predictive multiplicity & The disagreement in model predictions from a set of nearly equivalent accuracy models, or the Rashomon set of models~\cite{marx2020predictive}. The extent to which equivalent models reach conflicting predictions for the same instance. \\
Disagreement profile & The joint certainty status of an instance: whether the annotators and the models are each internally unanimous or divided, yielding the four profiles MC, MoD, HoD, and MD (Sec.~III-F). \\
Alignment & The degree to which model uncertainty or disagreement corresponds to human disagreement. How well model uncertainty or multiplicity reflects the disagreement observed among human annotators. \\
\bottomrule
\end{tabularx}
\end{table*}

% \begin{table*}[t]
% \centering
% \caption{Definitions of key terms used in this work.}
% \label{tab:key_terms}
% \begin{tabular}{p{0.15\textwidth} p{0.43\textwidth} p{0.25\textwidth}}
% \hline
% \textbf{Term} & \textbf{Definition} & \textbf{What it captures} \\
% \hline

% \textbf{Human Uncertainty} &
% Uncertainty arising from disagreement among human annotators when assigning labels to the same instance, as defined in \cite{mendes2025uncertainty,lan2025mind} &
% The inherent ambiguity or subjectivity of the ground truth label. \\

% \textbf{Model Uncertainty} &
% Uncertainty in a single model's prediction, typically quantified using predictive entropy or another uncertainty score derived from its predicted class probabilities. \cite{gawlikowski2023survey,hullermeier2021aleatoric} &
% How confident a model is about its prediction. \\

% \textbf{Predictive Multiplicity} &
% The disagreement in model predictions from a set of nearly equivalent accuracy models, or the Rashomon set of models \cite{marx2020predictive}. &
% The extent to which equivalent models reach conflicting predictions for the same instance. \\

% \textbf{Alignment} &
% The degree to which model uncertainty or disagreement corresponds to human uncertainty. &
% How well model uncertainty or multiplicity reflects the disagreement observed among human annotators. \\

% \hline
% \end{tabular}
% \end{table*}

\section{Related Work}

Uncertainty Quantification in deep neural networks has been studied extensively, especially in the last decade \cite{gal2016dropout,lakshminarayanan2017simple,gawlikowski2023survey,hullermeier2021aleatoric}. In these attempts, the general goal was to improve or calibrate the model uncertainty measures so that they align well with the model architecture, inductive bias and randomness. However, there is limited work in validating the alignment of this uncertainty against human uncertainty. Another powerful mechanism for measuring model safety in high stakes ML is predictive multiplicity, which measures the disagreement among equivalent models that are able to achieve nearly equivalent accuracy on a given dataset \cite{marx2020predictive}. 

We study a total of four predictive \textit{uncertainty} and \textit{multiplicity} metrics: single-model metrics (softmax confidence and predictive entropy), and predictive multiplicity (inter-model label disagreement and inter-model distribution divergence). These represent the most practical and widely-deployed approaches for uncertainty estimation in production systems. The motivation for using additional multiplicity metrics are that one can intuitively expect images where equivalent models disagree to also be where multiple humans disagree.

The work most relevant to us was by \cite{mendes2025uncertainty}, where the authors reported a weak correlation between human uncertainty and model uncertainty in classifying images from three benchmark vision datasets \texttt{CIFAR-10H}, \texttt{CIFARN} and \texttt{ImageNet}. 
However, their analysis uses one pre-trained model on each dataset, showing only limited possibilities of alignment. Our work extends this by considering the multiple models (8 in our study), from three different model families, extracting both uncertainty metrics of a single models and predictive multiplicity measures (across multiple equally good models), giving a richer analysis. 

 % Findings in \cite{mendes2025uncertainty} were similar to ours, but could not provide a deeper understanding of where the humans and models diverge in their perceptions. 

The two datasets (\texttt{FER+} and \texttt{CIFAR-10H}) used in this study were chosen due to the availability of multiple-human annotations. \texttt{CIFAR-10H} was created with the purpose of training models with soft labels for more robust classification \cite{peterson2019human}. This provided a novel method for creating models that learns the human disagreement at the training stage, leading to naturally more aligned models. In this study, we do not adopt the method of training with soft labels, but stick to fine-tuning pre-trained models with the hard labels that were taken from the majority vote of the human annotations. \texttt{FER+} was another dataset where the entire set including the training and validation instances contained multiple human annotations \cite{testoni2024asking}.

Our views align closely with authors in \cite{horchani2026beyond}, who argue that model uncertainty measures are not enough to find the most effective pool of samples in active learning. Poor calibration, noise and outliers in deep settings can misguide uncertainty, leading to poorly chosen difficult samples. This motivates approaches that detect ambiguous instances directly rather than inferring them from uncertainty, such as those in~\cite{cooper2024arbitrariness,singh2026robust}.

\section{Methods}

\subsection{Datasets}
In this work we use two image datasets, \texttt{FER+} \cite{barsoum2016training} and \texttt{CIFAR-10H}\cite{peterson2019human}, each of which has either the full or a part of the datapoints labelled by multiple human raters, thus enabling us to understand the degree of ambiguity humans have for a datapoint. A brief and relevant description of the datasets is given below.

\begin{itemize}
    \item \texttt{FER+} (Facial Expression Recognition)\cite{barsoum2016training}:  An extension of the original FER\cite{goodfellow2013challenges} dataset, \texttt{FER+} contains images of human faces ($48\times48$ greyscale) with different facial expressions. There are a total of $28,558$ training images, $3,579$ validation images, and $3,572$ test images. Each image can be associated with one of the $8$ class labels: \textit{neutral}, \textit{happiness}, \textit{surprise}, \textit{sadness}, \textit{anger}, \textit{disgust}, \textit{fear}, and \textit{contempt}. Every image is labeled by at least $3$, and at most $10$ human raters, each of whom assigns one of eight labels to an image. Therefore, each image will have a total of at most $10$ annotations, each indicating what facial expression the image contains according to the corresponding human rater. After removing annotators that failed the attention checks, each image was left with at most $10$ annotations.

    \item \texttt{CIFAR-10H}\cite{peterson2019human}: This dataset is an extension of the \texttt{CIFAR-10}\cite{krizhevsky2009learning} dataset. Original \texttt{CIFAR-10} dataset has $50,000$ training images and $10,000$ test images ($32\times32$, RGB). Each can be classified as one of the $10$ classes indicating what is present in the image: \textit{airplane}, \textit{automobile}, \textit{bird}, \textit{cat}, \textit{deer}, \textit{dog}, \textit{frog}, \textit{horse}, \textit{ship}, and \textit{truck}. The original \texttt{CIFAR-10} has one class label assigned to each image as the ground truth. The \texttt{CIFAR-10H} extension is identical in all aspects, but the $10,000$ test images are annotated by $51$ human raters assigning one of $10$ class  labels. The training and validation partitions of the dataset continue to have a single class label assigned to them.
    
\end{itemize}

\texttt{FER+} represents subjective task ambiguity (emotion is culturally  variable). \texttt{CIFAR-10H} represents objective task ambiguity (clear objects, but disagreement on boundary/low-quality cases). Together, they enable our alignment investigation across datasets that have ambiguity from two distinct sources.

\subsection{Models and Training}

To train models using the datasets, we have used a total of $8$ pre-trained vision models spanning three architectures: \texttt{ResNet} (18, 34, 50) \cite{he2016deep}, \texttt{EfficientNet} (B0, B1, B2) \cite{tan2019efficientnet}, and \texttt{MobileNetV3} (Small, Large) \cite{howard2019searching}. Each of these models is configured to work for input images of resolution $224\times224$ pixels. The dataset images were upsampled using bi-linear interpolation. We fine-tuned the models on the training split  (Adam\cite{kingma2014adam} optimiser, learning rate $10^{-4}$,  $10$ epochs, batch size 32). All models were accessed through the \textit{timm} \cite{wightman2021resnet} library in Python.

\texttt{CIFAR-10H} training partition has one ground truth label per image. All models were fine tuned using the training set with the available single class label as the target. For \texttt{FER+}, all the datapoints have class labels assigned by $10$ human raters. For our experiment, we chose the majority class label assigned to an image as the target labels for fine tuning. For example, if an image is labelled as \textit{surprise} by $7$ human raters and the remaining three assign the label \textit{fear}, then we chose the majority label \textit{surprise} as the target label for the fine tuning of the models.
 % REMOVING THIS: "We added a dropout layer ($p = 0.5$) before the final classifier to enable stochastic inference."
% TODO: Mention that the objective is to train each model so that they are in a decent predictive performance level. ALSO WE SHOULD CHECK IF OUR LEVELS OF ACCURACY IS ACCEPTABLE, OTHERWISE PEOPLE CAN SAY BECAUSE THE MODELS ARE WEAK (IF THE ACCURACY IS LOWER), THEN THE MODELS ARE MAYVE AGREEING MORE OR WHATEVER ARGUMENT.

\subsection{Human Disagreement Metric}
For each image in the dataset, we are provided with multiple human annotations corresponding to the categories that annotators believe best represent the image. We interpret the proportion of votes assigned to each category as its empirical vote probability. Human disagreement is quantified using the entropy of this vote distribution:

% \begin{equation}
% H_{\text{human}} = -\sum_{i=1}^{K} p_i \log p_i,
% \end{equation}

\begin{equation}
    H^{\mathrm{norm}}_{\mathrm{human}} = -\frac{1}{\log K}\sum_{i=1}^{K} p_i \log p_i
\end{equation}

where $K$ denotes the number of possible class labels and $p_i$ is the proportion of annotators who assign label $i$ to the image. $H^{\mathrm{norm}}_{\mathrm{human}} \in [0,1]$; higher values of $H_{\text{human}}^{\text{norm}}$ indicate greater disagreement among annotators, whereas $H_{\text{human}}^{\text{norm}} = 0$ corresponds to unanimous agreement. 

% To obtain a normalized measure of human uncertainty on a scale from 0 to 1, we divide the entropy for each image by the maximum entropy observed across the dataset, thus.

\subsection{Model Uncertainty and Disagreement Metrics}\label{subsec:model_un_dis}
We measure model uncertainty based on the probability distribution assigned by each model across the available class labels. Unlike the human annotations, which provide discrete votes from multiple annotators, a model produces a probability distribution over the $K$ possible classes for each image. Let $q_i$ denote the predicted probability assigned by the model to class $i$, where $\sum_{i=1}^{K} q_i = 1$.

For each model, we compute two measures of predictive uncertainty:

\begin{itemize}
    \item \textit{Softmax Confidence}: the maximum predicted class probability,
    \begin{equation}
    C_{\text{model}} = \max_{i} q_i.
    \end{equation}
    Higher values indicate that the model is more confident in its prediction.

    \item \textit{Predictive Entropy}: the entropy of the model's predicted probability distribution,
    % \begin{equation}
    % H_{\text{model}} = -\sum_{i=1}^{K} q_i \log q_i.
    % \end{equation}
    \begin{equation}
    H^{\mathrm{norm}}_{\mathrm{model}} = -\frac{1}{\log K}\sum_{i=1}^{K} q_i \log q_i
    \end{equation}
    
    $H^{\mathrm{norm}}_{\mathrm{model}} \in [0,1]$, where values closer to $0$ indicate greater model certainty and values closer to $1$ indicate relatively higher predictive uncertainty.
\end{itemize}

% \textbf{Stochastic Uncertainty (Monte Carlo Dropout):}
% We enable dropout during inference and perform 30 forward passes, computing entropy of the mean predicted distribution. Note: On pretrained models with fine-tuning in final layers only, MC dropout induces low stochasticity, resulting in nearly stable predictions across samples.

\subsection{Predictive Multiplicity}\label{subsec:predmul}
Using all 8 models as a subset of the Rashomon set of models for this image classification problem, we compute:
\begin{itemize}
    \item Inter-Model Label Disagreement
    For each image, we treat the predicted class from each model as a vote. We use the proportion of models predicting each category as the vote probability and calculate the entropy across this distribution:

    % \begin{equation}
    % H_{\text{model}} = -\sum_{i=1}^{K} r_i \log r_i,
    % \end{equation}

    \begin{equation}
        H^{\mathrm{norm}}_{\mathrm{vote}} = -\frac{1}{\log K}\sum_{i=1}^{K} r_i \log r_i
    \end{equation}

    $H_{\text{vote}}^{\text{norm}} \in [0,1]$, where values closer to $0$ indicate agreement among the models, while values close $1$ indicate high disagreement among the models.
    
    % where $c$ is the number of class labels and $r_i$ is the proportion of models that predict class $i$. Higher values of $H_{\text{model}}$ indicate greater disagreement among models, whereas $H_{\text{model}} = 0$ corresponds to unanimous model predictions.
    
    % To normalize the measure between 0 and 1, we divide by the maximum entropy observed across all images. Thus, $H_{\text{model}}^{\text{norm}} \in [0,1]$, where values closer to 0 indicate agreement among the models, while values close 1 indicate high disagreement among the models.
   
    \item \textit{Inter-Model Distribution Divergence}: We measure the divergence between model prediction distributions using the Jensen-Shannon divergence \cite{lin1991divergence}. Let $l_m$ denote the predictive probability distribution of model $m\in \mathcal{M}$, and let $\bar{l}$ denote the mean predictive distribution across all $\mathcal{M}$ models:

    \begin{equation}
    \bar{l} = \frac{1}{|\mathcal{M}|}\sum_{m\in\mathcal{M}} l_m
    \end{equation}
    
    The inter-model distribution divergence is then defined as:
    
    \begin{equation}
    D_{\mathrm{JSD}}
    =
    H(\bar{l})
    -
    \frac{1}{|\mathcal{M}|}\sum_{m\in\mathcal{M}} H(l_m),
    \end{equation}
    
    where $H(\cdot)$ denotes Shannon entropy. Higher values indicate a greater disagreement between the models' predictive distributions.

    \item \textbf{Model Family Disagreement:}
    We use models from different model families (\texttt{ResNet}, \texttt{EfficientNet} and \texttt{MobileNetV3}), and for each type, we use different versions (eg. \texttt{MobileNetV3} Small and Large), it would also be interesting to observe the disagreement within-family as well as between-family.

    Let $\mathcal{F}$ denote the set of model families $\mathcal{M}_f$ the models belonging to family $f$, $\mathcal{M}$ the full set of models, and $L_{m,i}$ the predictive distribution of model $m$ for image $i$. Let $\bar{L}_{f,i}$ denote the mean prediction of family $f$, and $\bar{L}_i$ the overall mean prediction. We then decompose ensemble disagreement into within-family and between-family components. For each model family $f$ (\texttt{ResNet}, \texttt{EfficientNet}, \texttt{MobileNetV3}), we compute:
    
    \textbf{Within-family disagreement} is averaged equally across all models:
    \begin{equation}
    D_{\text{within},i}
    =
    \frac{1}{|\mathcal{M}|}
    \sum_{f \in \mathcal{F}}
    \sum_{m \in \mathcal{M}_f}
    d_{\mathrm{JS}}
    \left(L_{m,i}\,\middle\|\,\bar{L}_{f,i}\right).
    \end{equation}
    
    \textbf{Between-family disagreement} gives equal weight to each family:
    \begin{equation}
    D_{\text{between},i}
    =
    \frac{1}{|\mathcal{F}|}
    \sum_{f \in \mathcal{F}}
    d_{\mathrm{JS}}
    \left(\bar{L}_{f,i}\,\middle\|\,\bar{L}_i\right).
    \end{equation}
    
    % TODO: mention if high is better or whatever.
    The \textbf{disagreement ratio} $R_i = D_{\text{between},i} / (D_{\text{within},i} + \epsilon)$ indicates whether the disagreement stems from architectural differences (ratio $>$ 1) or model-variant differences within architectures (ratio $<$ 1). $R_i>1$ indicates greater between-family disagreement and $R_i<1$ greater within-family disagreement.

\end{itemize}

\subsection{Alignment Analysis}\label{subsec:alin_ana}
Broadly, we compare model-side metrics to the human disagreement on two levels. First, using a discrete label-level analysis, where humans and models are considered aligned when their certainty status matches, that is, both are unanimous or both divide, and misaligned when one side is unanimous while the other divides. Second, using a continuous score-level correlation analysis, where each model-side uncertainty and multiplicity metric is correlated against human disagreement.

% \textbf{Discrete Label-Level}: We cross-tabulate human and model agreement on  hard labels to identify four quadrants:

% \begin{table}[H]
% \centering
% \caption{Human-model agreement quadrants.}
% \label{tab:quadrants}
% \begin{tabular}{lcc}
% \hline
%  & \textbf{Models Agree} & \textbf{Models Disagree} \\
% \hline
% \textbf{Humans Agree}    & \textbf{Q1} & \textbf{Q2} \\
% \textbf{Humans Disagree} & \textbf{Q3} & \textbf{Q4} \\
% \hline
% \end{tabular}
% \end{table}

% \textit{Q1} contains instances where humans and models reach unanimous agreement on the same label, representing the clearest cases for autonomous deployment. \textit{Q4} contains instances where both humans and models disagree, suggesting that model predictions should be deferred to human experts. \textit{Q2} and \textit{Q3} require further consideration. We focus on \textit{Q3}, where humans disagree but models produce unanimous predictions. These cases are particularly important because model confidence may not reflect the inherent ambiguity of the sample, limiting the trustworthiness of model uncertainty estimates.

\textbf{Discrete Label-Level:} We categorise each test instance by whether the human annotations and the model predictions are each internally unanimous, yielding four disagreement profiles: \textit{mutual consensus} (MC), where both the annotators and the models are unanimous; \textit{model-only disagreement} (MoD), where annotators are unanimous but the models divide; \textit{human-only disagreement} (HoD), where annotators divide but the models are unanimous; and \textit{mutual disagreement} (MD), where both divide. MC instances are the clearest candidates for autonomous deployment, provided the two unanimous labels coincide. MD instances are flagged as difficult by both parties, and deferring them to human experts is the natural policy. MoD instances trigger review on samples that humans resolve easily, that is, a cost in efficiency rather than safety. Our focus is on HoD: here the models are unanimously confident precisely where human annotators divide, so the ambiguity of the instance is masked by model agreement, and no confidence-based deployment rule can surface it. These are the cases that most directly limit the trustworthiness of model disagreement as a deployment signal.

% \textbf{Continuous Score-Level}: We compute Spearman rank correlation ($\rho$)  between each single model's uncertainty scores, aggregate model multiplicity metrics and human disagreement. Higher correlation indicates better alignment.
\textbf{Continuous Score-Level:} In addition to the multiplicity measures of Sec.~\ref{subsec:predmul}, we aggregate the single-model measures of Sec.~\ref{subsec:model_un_dis} across the $|\mathcal{M}|$ models: the average prediction score $\bar{C} = \frac{1}{|\mathcal{M}|}\sum_{m=1}^{|\mathcal{M}|} C_m$ and the average predictive entropy $\bar{H} = \frac{1}{|\mathcal{M}|}\sum_{m=1}^{|\mathcal{M}|} H^{\mathrm{norm}}_{m}$. We compute the Spearman rank correlation ($\rho$) between each single model's uncertainty scores, the aggregate measures, and human disagreement $H^{\mathrm{norm}}_{\mathrm{human}}$. Higher correlation indicates better alignment.

\section{Results}

The test accuracy of all 8 models after fine tuning is shown in Table \ref{tab:model_accuracy}. The baseline accuracy for models trained on \texttt{CIFAR-10} is approximately $85\%$ \cite{peterson2019human}, and for \texttt{FER} is $73\%$ \cite{khaireddin2021facial}. Our fine-tuned pre-trained models remain comparable to this range. The models on average are more accurate in the \texttt{CIFAR-10H} test instances as compared to \texttt{FER+}. \texttt{FER+} has training and validation samples that are labeled using majority human votes, that results in models that are fine tuned based on the majority votes, rather than the original training labels in the \texttt{FER} dataset that was not obtained by majority votes among independent human raters. This is not the case for \texttt{CIFAR-10H}, where the models are fine-tuned from train and validation labels that were \textit{not} selected from majority votes from multiple independent annotators, while the test labels were. 

\begin{table}[t]
\caption{Test accuracy (\%) of the 8 fine-tuned models (three families) on \texttt{CIFAR-10H} and \texttt{FER+}.}
\label{tab:model_accuracy}
\centering
\scriptsize
\setlength{\tabcolsep}{3pt}
\begin{tabular}{l ccc ccc cc}
\toprule
 & \multicolumn{3}{c}{\texttt{ResNet}} & \multicolumn{3}{c}{\texttt{EfficientNet}} & \multicolumn{2}{c}{\texttt{MobileNetV3}} \\
\cmidrule(lr){2-4} \cmidrule(lr){5-7} \cmidrule(lr){8-9}
 & 18 & 34 & 50 & B0 & B1 & B2 & Small & Large \\
\midrule
\texttt{CIFAR-10H} & 94.34 & 96.31 & 96.05 & 96.13 & 94.96 & 96.20 & 92.95 & 95.10 \\
\texttt{FER+}      & 77.38 & 77.58 & 78.44 & 78.28 & 80.32 & 77.41 & 77.86 & 78.72 \\
\bottomrule
\end{tabular}
\end{table}

% \begin{table}
% \centering
% \caption{Test accuracy (\%) of the 8 pre-trained models after fine-tuning on \texttt{CIFAR-10H} and \texttt{FER+}.}
% \label{tab:model_accuracy}
% \begin{tabular}{lcc}
% \hline
% \textbf{Model} & \textbf{CIFAR-10H} & FER+ \\
% \hline
% ResNet-18           & 94.34 & 77.38 \\
% ResNet-34           & 96.31 & 77.58 \\
% ResNet-50           & 96.05 & 78.44 \\
% EfficientNet-B0     & 96.13 & 78.28 \\
% EfficientNet-B1     & 94.96 & 80.32 \\
% EfficientNet-B2     & 96.20 & 77.41 \\
% MobileNetV3-Small   & 92.95 & 77.86 \\
% MobileNetV3-Large   & 95.10 & 78.72 \\
% \hline
% \end{tabular}
% \end{table}

 \begin{figure*}[h]
    \centering
    \begin{subfigure}{0.46\textwidth}
        \centering
        \includegraphics[width=\linewidth]{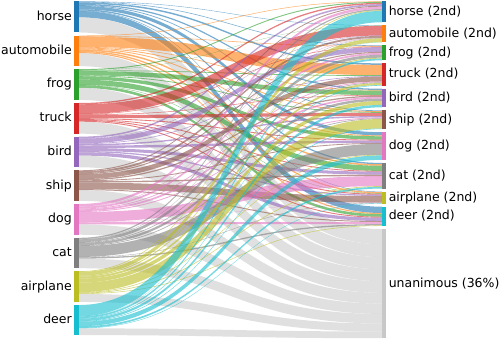}
        \caption{\texttt{CIFAR-10H}: Majority and Second Majority Human Annotations}
        \label{fig:cifar_model_vs_human}
    \end{subfigure}
    \hfill
    \begin{subfigure}{0.46\textwidth}
        \centering
        \includegraphics[width=\linewidth]{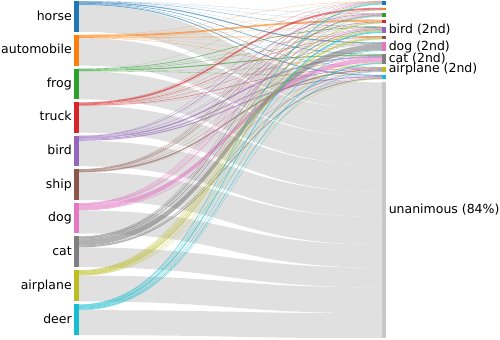}
        \caption{\texttt{CIFAR-10H}: Majority and Second Majority Model Predictions}
        \label{fig:second_majorrt}
    \end{subfigure}
    \caption{Overview of test images in \texttt{CIFAR-10H} and their voted majority class on the left, and the second majority on the right. Images that had a unanimous vote are denoted in grey.}
    \label{fig:alluvial_cifar}
\end{figure*}

 \begin{figure*}[h]
    \centering
    \begin{subfigure}{0.46\textwidth}
        \centering
        \includegraphics[width=\linewidth]{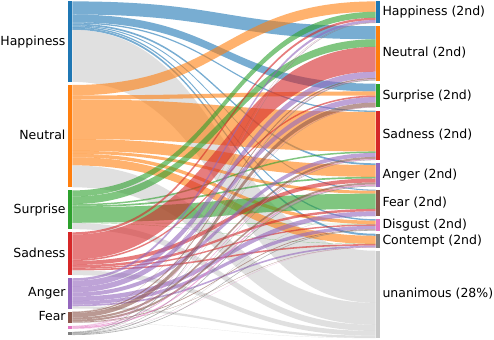}
        \caption{\texttt{FER+}: Majority and Second Majority Human Annotations}
        \label{fig:cifar_model_vs_human}
    \end{subfigure}
    \hfill
    \begin{subfigure}{0.46\textwidth}
        \centering
        \includegraphics[width=\linewidth]{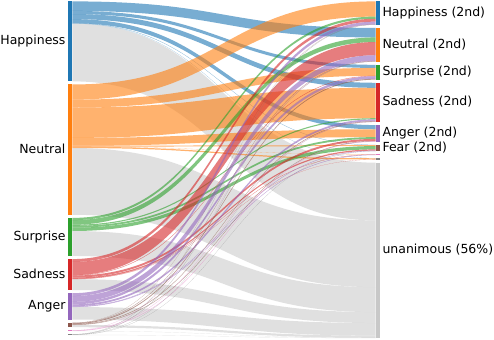}
        \caption{\texttt{FER+}: Majority and Second Majority Model Predictions}
        \label{fig:second_majorrt}
    \end{subfigure}
    \caption{Overview of test images in \texttt{FER+} and their voted majority class on the left, and the second majority on the right. Images that had a unanimous vote are denoted in grey.}
    \label{fig:alluvial_fer}
\end{figure*}

% We observe a clear difference between the pattern of human annotations and model predictions, as shown in Fig. \ref{fig:majority_vs_second_majority_FER+} and Fig. \ref{fig:majority_vs_second_majority_CIFAR10h}. In \texttt{CIFAR-10H}, models tend to make more unanimous predictions, whereas human annotators more frequently assign a second-majority vote across images. For example, the category 'deer' has majority of images that are also voted into a second majority category of 'horse' by human annotators, while the models predict majority of the 'deer' images unanimously. In \texttt{FER+}, human annotations are highly contentious across all emotions, a pattern that is also reflected in the model predictions. Overall, the level of disagreement between models and humans appears to be more closely aligned in \texttt{FER+} than in \texttt{CIFAR-10H}. This is even more clearly shown in Fig. \ref{fig:categories_human_vs_model} where each model and human disagreement in each class is directly compared. In \texttt{FER+}, the bars align quite closely, but in \texttt{CIFAR-10H}, there is no clear trend or equivalence between the model-human disagreement in each class. 

Figs.~\ref{fig:alluvial_cifar} and~\ref{fig:alluvial_fer} show the flow from majority to second-majority labels for human annotations and model predictions. Unanimously labelled instances are routed to the grey section, and classes are ordered identically in both panels to allow direct comparison. The asymmetry is clear. On \texttt{CIFAR-10H}, the models are unanimous on $83.5\%$ of test images against $35.7\%$ for the annotators. Whereas, on \texttt{FER+}, $56.3\%$ against $28.0\%$. Where human annotators divide, their second-majority votes fall in semantically related pairs. For example \textit{deer} with \textit{horse} and \textit{cat} with \textit{dog} on CIFAR-10H. While in \texttt{FER+}, \textit{sadness} and \textit{contempt} with \textit{neutral} on FER+. Human disagreement therefore reflects structured class confusability. On \texttt{FER+} the models' divided predictions fall largely on the same emotion pairs, and the shares of contested instances are closer, consistent with the stronger score-level alignment reported in Sec.~\ref{subsec:cont_score_align}. The excess grey in each model panel over its human panel is the visual counterpart of the HoD share in Tables~\ref{tab:human_model_agreement_fer} and~\ref{tab:cm_cifar}: instances on which model unanimity conceals a divided human vote.

\subsection{Discrete Label-Level Alignment}

\begin{table}[h!]
\caption{Human-model agreement on the \texttt{FER+} test set
($n{=}3572$). Cells show counts with percentages of the test set in parentheses.}
\label{tab:human_model_agreement_fer}
\centering
\scriptsize
\setlength{\tabcolsep}{4pt}
\begin{tabular}{l| ll |l}
\toprule
 & Models Agree & Models Disagree & Total \\
\midrule
Humans Agree    &  MC: $815$ ($22.8$\%) &  MoD: $186$ ($5.2$\%)  & $1001$ ($28.0$\%) \\
Humans Disagree & HoD: $1195$ ($33.5$\%) & MD: $1376$ ($38.5$\%) & $2571$ ($72.0$\%) \\
\midrule
Total           & $2010$ ($56.3$\%) & $1562$ ($43.7$\%) & $3572$ \\
\bottomrule
\end{tabular}
\end{table}

From Table \ref{tab:human_model_agreement_fer}, the \textbf{human-model-alignment} on the \texttt{FER+} test instances is \textbf{61.34\%} (MC + MD). Among the remaining 38.66\% of images, where the models and humans do not align, there are $1195$ images that were predicted to belong to same class by all 8 models, while humans still expressed some level of disagreement. This is a concerning number of images that would typically be automatically deployed since the measure of model agreement would flag these as stable, unambiguous images. Among these $1195$ images, $1092$ were still classified into the correct true category, which would be masked by measures that only measure whether the prediction is correct. However, $103$ of these $1195$ images were still misclassified by majority of the models. For the images that the humans and models agreed to be ambiguous or assigned more than one label, $879$ images were correctly predicted, while $497$ were incorrectly predicted by a majority of the models. This is another indication of how despite both models and humans agree that the image is ambiguous, the underlying perception still diverges, as they do not align on the majority label for that image.

\begin{table}[h!]
\caption{Human--model agreement on the \texttt{CIFAR-10H} test set
($n{=}10{,}000$). Cells show counts with percentages of the
test set in parentheses.}
\label{tab:cm_cifar}
\centering
\scriptsize
\setlength{\tabcolsep}{4pt}
\begin{tabular}{l| ll |l}
\toprule
 & Models Agree & Models Disagree & Total \\
\midrule
Humans Agree    & MC: $3310$ ($33.1$\%) &  MoD: $255$ ($2.6$\%)  &  $3565$ (35.7\%) \\
Humans Disagree & HoD: $5044$ ($50.4$\%) & MD: $1391$ ($13.9$\%) &  $6435$ (64.4\%) \\
\midrule
Total           & $8354$ ($83.5$\%) & $1646$ ($16.5$\%) & $10,000$ \\
\bottomrule
\end{tabular}
\end{table}

From table \ref{tab:cm_cifar}, the \textbf{human-model-alignment} on the \texttt{CIFAR-10H} dataset is \textbf{47.01\%}. This is much lower than the alignment reported in \texttt{FER+} dataset, despite the higher accuracy of the models trained on \texttt{CIFAR-10H} as seen in Table \ref{tab:model_accuracy}. In this confusion matrix for the \texttt{CIFAR-10H} dataset there are $5044$ images that the models agree on, while human annotators still express contention. 8 out of these $5044$ images were misclassified by majority of the models, while the rest were correctly classified. 

\begin{figure*}[h]
  \centering
  \includegraphics[width=\textwidth]{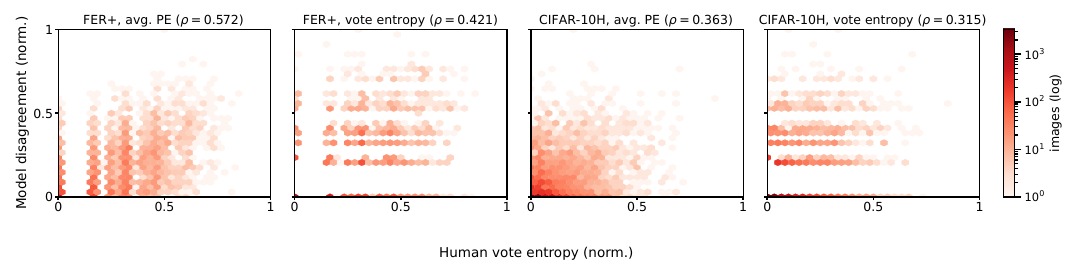}
  \caption{Per-instance human disagreement ($H^{\mathrm{norm}}_{\mathrm{human}}$) vs.\ model disagreement for each dataset and model-side measure; colour gives image counts on a log scale. Even in the best-aligned setting (\texttt{FER+}, average predictive entropy ($\bar{H}$), $\rho=0.572$), instances are widely dispersed off the diagonal. On \texttt{CIFAR-10H} the weak alignment holds for the continuous measure as well ($\rho=0.363$). For inter-model vote entropy ($H^{\mathrm{norm}}_{\mathrm{vote}}$, $\rho=0.315$), the dense band at zero, which is the $83.5\%$ of images on which all eight models agree, spans the full range of human disagreement.}
  \label{fig:scatter}
\end{figure*}

\subsection{Continuous Score-Level Alignment}\label{subsec:cont_score_align}

Fig.~\ref{fig:scatter} shows the joint distribution of human and model disagreement per instance. Even in the best-aligned setting (FER+, average predictive entropy), the mass is widely dispersed rather than concentrated on the diagonal. On \texttt{CIFAR-10H}, the dense band at zero model vote entropy spans the full range of human disagreement. These are the HoD instances of Table~\ref{tab:cm_cifar} observed at score level.

\begin{figure*}[h!]
    \centering
    \begin{subfigure}{0.49\linewidth}
        \centering
        \includegraphics[width=\linewidth]{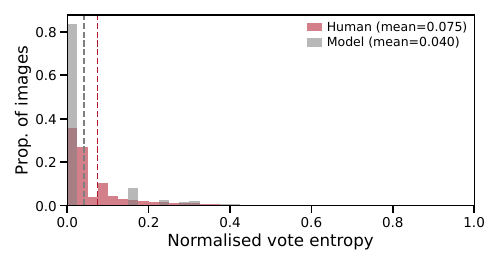}
        \caption{\texttt{CIFAR-10H}}
        \label{fig:CIFAR_Dist}
    \end{subfigure}
    \begin{subfigure}{0.49\linewidth}
        \centering
        \includegraphics[width=\linewidth]{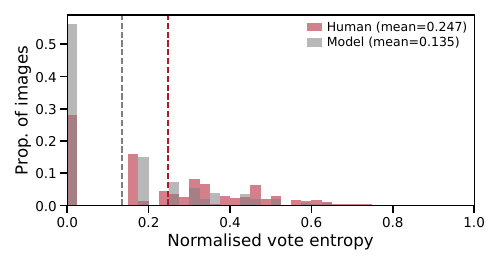}
        \caption{\texttt{FER+}}
        \label{fig:FER_Dist}
    \end{subfigure}
    \caption{Distributions of normalised vote entropy for human annotations ($H^{\mathrm{norm}}_{\mathrm{human}}$) and model predictions ($H^{\mathrm{norm}}_{\mathrm{vote}}$). Dashed lines mark mean values.}
    \label{fig:human_model_disagreement}
\end{figure*}

\begin{figure*}[h!]
    \centering
    \begin{subfigure}{0.49\textwidth}
        \centering
        \includegraphics[width=\linewidth]{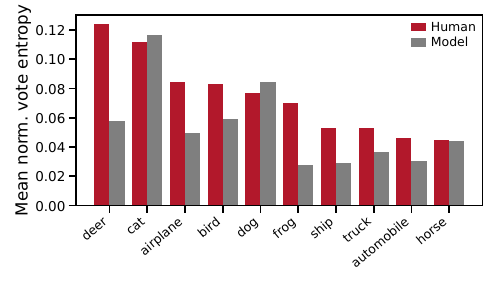}
        \caption{\texttt{CIFAR-10H}}
        \label{fig:cifar_model_vs_human}
    \end{subfigure}
    % \hfill
    \begin{subfigure}{0.49\textwidth}
        \centering
        \includegraphics[width=\linewidth]{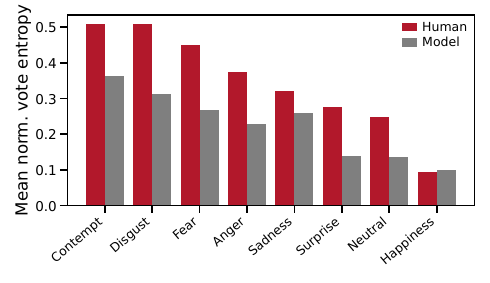}
        \caption{\texttt{FER+}}
        \label{fig:fer_model_vs_human}
    \end{subfigure}
    % \caption{Model vs.\ human normalised vote entropy across binned average prediction scores from the eight trained models.}
    \caption{Mean per-class human disagreement ($H^{\mathrm{norm}}_{\mathrm{human}}$) and model disagreement ($H^{\mathrm{norm}}_{\mathrm{vote}}$) across classes, sorted by human disagreement.}
    \label{fig:categories_human_vs_model}
\end{figure*}

The human annotators show more disagreement in the \texttt{FER+} dataset as compared to the \texttt{CIFAR-10H} dataset, see Fig.~\ref{fig:human_model_disagreement}. The average vote entropy among humans in \texttt{FER+} is $0.247$, while that in \texttt{CIFAR-10H} is $0.075$. Majority of the images in \texttt{CIFAR-10H} with some level of disagreement among the humans fall well below the average vote entropy, indicating that most of the ambiguous images have relatively lower levels of contention. In \texttt{FER+}, the images with disagreement remain spread across different values, and some images are seen to have vote entropies as high as $0.65$. In other words, the images in \texttt{FER+} are extremely confusing, leading to higher vote entropy than the images in \texttt{CIFAR-10H} which are less confusing, and have lower vote entropy values.

The per-class breakdown in Fig.~\ref{fig:categories_human_vs_model} shows where this disagreement concentrates: \emph{contempt}, \emph{disgust}, and \emph{fear} on FER+, and \emph{deer} and \emph{cat} on \texttt{CIFAR-10H}. Model disagreement, while consistently lower, follows the same per-class ordering.

% We dig deeper into the alignment by measuring the correlation between human and model alignment by using \textit{continuous} reliability scores that are measured on the model-side as well as the human-side.
We examine the alignment further by correlating \textit{continuous} reliability scores on the model side with human disagreement.
From Table \ref{tab:single_model_corr} it is evident that the single-model level metrics align weakly in \texttt{CIFAR-10H} ($0.24$ - $0.35$), and moderately in \texttt{FER+} ($0.41$ - $0.55$). Single model level uncertainty metrics reflect reliability of one model alone, and thus cannot capture the diversity of human votes that were made to a given image, and this result supports our claim.

\begin{table}[t]
\caption{Spearman correlations ($\rho$) between single-model uncertainty measures and normalised human vote entropy on \texttt{FER+} and
\texttt{CIFAR-10H}. All correlations are statistically significant ($p<0.05$).
SC: Softmax Confidence; PE: Predictive Entropy.}
\label{tab:single_model_corr}
\centering
\scriptsize
\setlength{\tabcolsep}{3pt}
\begin{tabular}{ll rrr rrr rr}
\toprule
 & & \multicolumn{3}{c}{\texttt{ResNet}} & \multicolumn{3}{c}{\texttt{EfficientNet}} & \multicolumn{2}{c}{\texttt{MobileNetV3}} \\
\cmidrule(lr){3-5} \cmidrule(lr){6-8} \cmidrule(lr){9-10}
 & & 18 & 34 & 50 & B0 & B1 & B2 & Small & Large \\
\midrule
\multirow{2}{*}{\texttt{CIFAR-10H}} & SC & -0.276 & -0.237 & -0.243 & -0.340 & -0.344 & -0.317 & -0.344 & -0.350 \\
                           & PE &  0.275 &  0.235 &  0.241 &  0.344 &  0.346 &  0.327 &  0.346 &  0.347 \\
\midrule
\multirow{2}{*}{\texttt{FER+}}      & SC & -0.470 & -0.450 & -0.420 & -0.438 & -0.550 & -0.413 & -0.488 & -0.451 \\
                           & PE &  0.474 &  0.450 &  0.421 &  0.442 &  0.554 &  0.416 &  0.491 &  0.453 \\
\bottomrule
\end{tabular}
\end{table}

% \begin{table}[ht]
% \centering
% \caption{Spearman rank correlations ($\rho$) between predictive multiplicity measures and normalised human vote entropy on the \texttt{FER+} and \texttt{CIFAR-10H} datasets.}
% \label{tab:aggregate_model_correlations}
% \begin{tabular}{lcc}
% \hline
% \textbf{Multiplicity Metrics} & \texttt{FER+} ($\rho$) & \textbf{\texttt{CIFAR-10H} ($\rho$)} \\
% \hline
% Average Prediction Score & -0.561 & -0.360 \\
% % Average MCU               &  0.572 &  0.363 \\
% Average PE                &  0.572 &  0.363 \\
% Inter-Model Vote Entropy  &  0.421 &  0.006 \\
% Within-Family Disagreement  &  0.523 &  0.201 \\
% Between-Family Disagreement &  0.471 &  0.174 \\
% Disagreement Ratio           & -0.314 & -0.184 \\
% \hline
% \end{tabular}
% \end{table}
\begin{table}[t]
\caption{Spearman correlations ($\rho$) between the aggregate model-side
measures (Secs.~\ref{subsec:model_un_dis}-\ref{subsec:alin_ana}) and human disagreement
$H^{\mathrm{norm}}_{\mathrm{human}}$. All correlations are statistically
significant ($p<0.05$).}
\label{tab:aggregate_corr}
\centering
\scriptsize
\setlength{\tabcolsep}{3.8pt}
\resizebox{\columnwidth}{!}{
\begin{tabular}{l ccccccc}
\toprule
 & $\bar{C}$ & $\bar{H}$ & $H^{\mathrm{norm}}_{\mathrm{vote}}$ & $D_{\mathrm{JSD}}$
 & $D_{\mathrm{within}}$ & $D_{\mathrm{between}}$ & $R$\\
\midrule
CIFAR-10H & $-0.360$ & $0.363$ & $0.315$ & $0.184$ & $0.201$ & $0.174$ & $-0.184$ \\
FER+      & $-0.561$ & $0.572$ & $0.421$ & $0.507$ & $0.523$ & $0.471$ & $-0.314$ \\
\bottomrule
\end{tabular}}
\end{table}
We also consider the aggregate model uncertainty metrics as shown in Table \ref{tab:aggregate_corr}. Here, the correlation between human disagreement and model disagreement is relatively better in \texttt{FER+} as compared to \texttt{CIFAR-10H}, however, in both cases this alignment remains weak to moderate. On \texttt{CIFAR-10H}, inter-model vote entropy aligns at $\rho = 0.315$, slightly below the continuous measures ($0.363$): being discrete, it is exactly zero on the $83.5\%$ of images where all eight models agree, and carries information only through the remaining $16.5\%$. More importantly, unanimity itself is uninformative about ambiguity, that is, the models are unanimous on $93\%$ of images where humans agree, but also on $78\%$ of images where they disagree. Model agreement, whether measured discretely or continuously, does not separate contested from uncontested instances.

A confidence threshold does little to remove this ambiguity. Above an average prediction score of $0.95$, covering $49.1\%$ of the \texttt{FER+} and $84.0\%$ of the \texttt{CIFAR-10H} test instances, model disagreement is nearly eliminated ($0.015$ and $0.009$ mean normalised vote entropy), yet mean human disagreement remains at $0.156$ and $0.057$, several times higher. The instances annotators found ambiguous largely end up above the threshold, where they would be decided automatically.

\section{Conclusion}

% SAME THING BUT SOME EDITS

% Conect with the introduction
Does model uncertainty track human ambiguity? On the evidence of two multi-annotator vision benchmarks, largely not. Our hypothesis held that instances which divide human annotators should be instances on which models disagree. This fails on both datasets: among the instances on which annotators divided, the eight models produced conflicting predictions on only $21.6\%$ (\texttt{CIFAR-10H}) and $53.5\%$ (\texttt{FER+}). On the remainder, unanimous predictions concealed a divided human vote. At the score level, single-model uncertainty correlates only weakly to moderately with human disagreement ($\rho = 0.24$--$0.55$), predictive multiplicity offers modest improvement, and a confidence threshold retains most of the instances humans found ambiguous.

The misalignment follows a consistent pattern. Alignment is stronger on \texttt{FER+}, whose fine-tuning labels derive from majority votes over multiple annotators, than on \texttt{CIFAR-10H}, whose labels do not, and human disagreement concentrates in semantically related class pairs that the models resolve unanimously. Hard-label training gives models no signal about annotation ambiguity, so the confident unanimity we observe is the expected product of standard practice rather than an anomaly.

A natural objection is that, when humans and models agree on a label, misaligned uncertainty estimates do not matter. This holds only where the ground truth is known. In deployment, predictions are effectively unsupervised: uncertainty estimates guide decisions precisely where the truth is yet to be uncovered, and an estimate weakly aligned with human judgement can then be harmful. Model uncertainty remains a valuable signal for improving model performance~\cite{hsu2024dropout}. When used to administer decisions that affect human lives, such as selective classification, active learning, and clinical decision support, it should not be presumed inherently trustworthy.

For practitioners, the implication is direct: the alignment between model-side measures and human disagreement should be verified before those measures are relied upon, which requires evaluation sets with multiple annotations per instance. Benchmark suites, including those for scientific AI, should therefore incorporate multi-annotator evaluation sets, so that uncertainty can be validated against human disagreement rather than accuracy against a single label alone. This work can be extended with further uncertainty measures, such as Monte Carlo dropout~\cite{gal2016dropout}, deep ensemble disagreement~\cite{lakshminarayanan2017simple}, and predictive multiplicity from Rashomon sets of lower tolerance; with per-annotator confidence scores, so that ambiguity can be studied at the level of individual raters; and with model disagreement tracked across training epochs to guide early decisions about model tuning.

\bibliographystyle{ieeetr}
\bibliography{references}

@inproceedings{hsu2024dropout,
  title={Dropout-based rashomon set exploration for efficient predictive multiplicity estimation},
  author={Hsu, Hsiang and Li, Guihong and Hu, Shaohan and Chen, Chun-Fu},
  booktitle={International Conference on Learning Representations},
  volume={2024},
  pages={33104--33144},
  year={2024}
}

@inproceedings{marx2020predictive,
  title={Predictive multiplicity in classification},
  author={Marx, Charles and Calmon, Flavio and Ustun, Berk},
  booktitle={International conference on machine learning},
  pages={6765--6774},
  year={2020},
  organization={PMLR}
}

@article{lakshminarayanan2017simple,
  title={Simple and scalable predictive uncertainty estimation using deep ensembles},
  author={Lakshminarayanan, Balaji and Pritzel, Alexander and Blundell, Charles},
  journal={Advances in neural information processing systems},
  volume={30},
  year={2017}
}

@inproceedings{gal2016dropout,
  title={Dropout as a bayesian approximation: Representing model uncertainty in deep learning},
  author={Gal, Yarin and Ghahramani, Zoubin},
  booktitle={international conference on machine learning},
  pages={1050--1059},
  year={2016},
  organization={PMLR}
}

@inproceedings{mendes2025uncertainty,
  title={Uncertainty estimation by human perception versus neural models},
  author={Mendes, Pedro and Romano, Paolo and Garlan, David},
  booktitle={Pacific Rim International Conference on Artificial Intelligence},
  pages={170--182},
  year={2025},
  organization={Springer}
}

@article{krizhevsky2009learning,
  title={Learning multiple layers of features from tiny images},
  author={Krizhevsky, Alex and Hinton, Geoffrey and others},
  year={2009},
  publisher={Toronto, ON, Canada}
}

@inproceedings{he2016deep,
  title={Deep residual learning for image recognition},
  author={He, Kaiming and Zhang, Xiangyu and Ren, Shaoqing and Sun, Jian},
  booktitle={Proceedings of the IEEE conference on computer vision and pattern recognition},
  pages={770--778},
  year={2016}
}

@inproceedings{howard2019searching,
  title={Searching for mobilenetv3},
  author={Howard, Andrew and Sandler, Mark and Chu, Grace and Chen, Liang-Chieh and Chen, Bo and Tan, Mingxing and Wang, Weijun and Zhu, Yukun and Pang, Ruoming and Vasudevan, Vijay and others},
  booktitle={Proceedings of the IEEE/CVF international conference on computer vision},
  pages={1314--1324},
  year={2019}
}

@inproceedings{tan2019efficientnet,
  title={Efficientnet: Rethinking model scaling for convolutional neural networks},
  author={Tan, Mingxing and Le, Quoc},
  booktitle={International conference on machine learning},
  pages={6105--6114},
  year={2019},
  organization={PmLR}
}

@techreport{settles2009active,
  title={Active Learning Literature Survey},
  author={Settles, Burr},
  institution={University of Wisconsin--Madison},
  number={1648},
  year={2009}
}

@article{lewis1994sequential,
  title={A sequential algorithm for training text classifiers},
  author={Lewis, David D and Gale, William A},
  journal={Proceedings of the 17th Annual International ACM SIGIR Conference on Research and Development in Information Retrieval},
  year={1994}
}

@article{geifman2017deep,
  title={Deep Active Learning for Biased Datasets via Fisher Kernel Self-Supervision},
  author={Geifman, Yonatan and El-Yaniv, Ran},
  journal={International Conference on Machine Learning},
  year={2017}
}

@inproceedings{geifman2019selectivenet,
  title={Selectivenet: A deep neural network with an integrated reject option},
  author={Geifman, Yonatan and El-Yaniv, Ran},
  booktitle={International conference on machine learning},
  pages={2151--2159},
  year={2019},
  organization={PMLR}
}

@inproceedings{testoni2024asking,
  title={Asking the right question at the right time: Human and model uncertainty guidance to ask clarification questions},
  author={Testoni, Alberto and Fern{\'a}ndez, Raquel},
  booktitle={Proceedings of the 18th Conference of the European Chapter of the Association for Computational Linguistics (Volume 1: Long Papers)},
  pages={258--275},
  year={2024}
}

@article{michelmore2018evaluating,
  title={Evaluating uncertainty quantification in end-to-end autonomous driving control},
  author={Michelmore, Rhiannon and Kwiatkowska, Marta and Gal, Yarin},
  journal={arXiv preprint arXiv:1811.06817},
  year={2018}
}

@article{hullermeier2021aleatoric,
  title={Aleatoric and epistemic uncertainty in machine learning: An introduction to concepts and methods},
  author={H{\"u}llermeier, Eyke and Waegeman, Willem},
  journal={Machine learning},
  volume={110},
  number={3},
  pages={457--506},
  year={2021},
  publisher={Springer}
}

@article{gawlikowski2023survey,
  title={A survey of uncertainty in deep neural networks: J. Gawlikowski et al.},
  author={Gawlikowski, Jakob and Tassi, Cedrique Rovile Njieutcheu and Ali, Mohsin and Lee, Jongseok and Humt, Matthias and Feng, Jianxiang and Kruspe, Anna and Triebel, Rudolph and Jung, Peter and Roscher, Ribana and others},
  journal={Artificial intelligence review},
  volume={56},
  number={Suppl 1},
  pages={1513--1589},
  year={2023},
  publisher={Springer}
}

@article{davani2022dealing,
  title={Dealing with disagreements: Looking beyond the majority vote in subjective annotations},
  author={Davani, Aida Mostafazadeh and D{\'\i}az, Mark and Prabhakaran, Vinodkumar},
  journal={Transactions of the Association for Computational Linguistics},
  volume={10},
  pages={92--110},
  year={2022},
  publisher={MIT Press One Rogers Street, Cambridge, MA 02142-1209, USA journals-info~…}
}

@inproceedings{lan2025mind,
  title={Mind the uncertainty in human disagreement: Evaluating discrepancies between model predictions and human responses in vqa},
  author={Lan, Jian and Frassinelli, Diego and Plank, Barbara},
  booktitle={Proceedings of the AAAI Conference on Artificial Intelligence},
  volume={39},
  number={4},
  pages={4446--4454},
  year={2025}
}

@article{pavlick-kwiatkowski-2019-inherent,
    title = "Inherent Disagreements in Human Textual Inferences",
    author = "Pavlick, Ellie  and
      Kwiatkowski, Tom",
    editor = "Lee, Lillian  and
      Johnson, Mark  and
      Roark, Brian  and
      Nenkova, Ani",
    journal = "Transactions of the Association for Computational Linguistics",
    volume = "7",
    year = "2019",
    address = "Cambridge, MA",
    publisher = "MIT Press",
    url = "https://aclanthology.org/Q19-1043/",
    doi = "10.1162/tacl_a_00293",
    pages = "677--694"
}

@article{lin1991divergence,
  title={Divergence measures based on the Shannon entropy},
  author={Lin, Jianhua},
  journal={IEEE Transactions on Information theory},
  volume={37},
  number={1},
  pages={145--151},
  year={1991},
  publisher={IEEE}
}

@article{kingma2014adam,
  title={Adam: A method for stochastic optimization},
  author={Kingma, Diederik P and Ba, Jimmy},
  journal={arXiv preprint arXiv:1412.6980},
  year={2014}
}

@article{wightman2021resnet,
  title={Resnet strikes back: An improved training procedure in timm},
  author={Wightman, Ross and Touvron, Hugo and J{\'e}gou, Herv{\'e}},
  journal={arXiv preprint arXiv:2110.00476},
  year={2021}
}

@article{khaireddin2021facial,
  title={Facial emotion recognition: State of the art performance on FER2013},
  author={Khaireddin, Yousif and Chen, Zhuofa},
  journal={arXiv preprint arXiv:2105.03588},
  year={2021}
}

@inproceedings{peterson2019human,
  title={Human uncertainty makes classification more robust},
  author={Peterson, Joshua C and Battleday, Ruairidh M and Griffiths, Thomas L and Russakovsky, Olga},
  booktitle={Proceedings of the IEEE/CVF international conference on computer vision},
  pages={9617--9626},
  year={2019}
}

@article{de2026uncertainty,
  title={Uncertainty quantification in machine learning for biosignal applications-a review},
  author={de Jong, Ivo Pascal and Sburlea, Andreea Ioana and Valdenegro-Toro, Matias},
  journal={Journal of Healthcare Informatics Research},
  pages={1--55},
  year={2026},
  publisher={Springer}
}

@article{horchani2026beyond,
  title={Beyond uncertainty in modern active learning for trustworthy AI},
  author={Horchani, Ridha},
  journal={Frontiers in Artificial Intelligence},
  volume={9},
  pages={1844765},
  year={2026},
  publisher={Frontiers Media SA}
}

@inproceedings{goodfellow2013challenges,
  title={Challenges in representation learning: A report on three machine learning contests},
  author={Goodfellow, Ian J and Erhan, Dumitru and Carrier, Pierre Luc and Courville, Aaron and Mirza, Mehdi and Hamner, Ben and Cukierski, Will and Tang, Yichuan and Thaler, David and Lee, Dong-Hyun and others},
  booktitle={International conference on neural information processing},
  pages={117--124},
  year={2013},
  organization={Springer}
}

@inproceedings{barsoum2016training,
  title={Training deep networks for facial expression recognition with crowd-sourced label distribution},
  author={Barsoum, Emad and Zhang, Cha and Ferrer, Cristian Canton and Zhang, Zhengyou},
  booktitle={Proceedings of the 18th ACM international conference on multimodal interaction},
  pages={279--283},
  year={2016}
}

@article{singh2026robust,
  title={Robust Ambiguity Detection ({RAD}) From Model-and Feature-Space Consistency},
  author={Singh, Manya and Keane, Mark T and Pakrashi, Arjun},
  journal={arXiv preprint arXiv:2608.11541},
  year={2026}
}

@inproceedings{cooper2024arbitrariness,
  title={Arbitrariness and social prediction: The confounding role of variance in fair classification},
  author={Cooper, A Feder and Lee, Katherine and Choksi, Madiha Zahrah and Barocas, Solon and De Sa, Christopher and Grimmelmann, James and Kleinberg, Jon and Sen, Siddhartha and Zhang, Baobao},
  booktitle={Proceedings of the AAAI Conference on Artificial Intelligence},
  volume={38},
  number={20},
  pages={22004--22012},
  year={2024}
}
\end{document}